\documentclass[10pt,twocolumn,letterpaper]{article}

\usepackage{iccv}              

\definecolor{iccvblue}{rgb}{0.21,0.49,0.74}
\usepackage[pagebackref,breaklinks,colorlinks,allcolors=iccvblue]{hyperref}
\usepackage{algorithmic}
\usepackage{algorithm}
\usepackage{amsmath}
\usepackage{multirow}
\usepackage{makecell}
\usepackage{bbding}
\usepackage{graphicx}

\def\paperID{7295} 
\def\confName{ICCV}
\def\confYear{2025}

\title{ExGes:  Expressive  Human Motion Retrieval and Modulation for\\ Audio-Driven Gesture Synthesis}

\author{Xukun Zhou\\
Renmin University\\
{\tt\small xukun\_zhou@ruc.edu.cn}
\and Fengxin Li\\
Renmin University\\
{\tt\small lifengxin@ruc.edu.cn}
\and Ming Chen\\
Kuaishou Technology\\
{\tt\small chenming09@kuaishou.com}
\and
Yan Zhou\\
Kuaishou Technology\\
{\tt\small zhouyan03@kuaishou.com}
\and 
Pengfei Wan\\
Kuaishou Technology\\
{\tt\small wanpengfei@kuaishou.com}
\and
Di Zhang\\
Kuaishou Technology\\
{\tt\small zhangdi08@kuaishou.com}
\and Yeying Jin \\
National University of Singapore\\
{\tt \small jinyeying@u.nus.edu}
\and Zhaoxin Fan\\
Renmin University\\
{\tt\small zhaoxinf@ruc.edu.cn}
\and Hongyan Liu\\
Tsinghua University\\
{\tt\small hyliu@tsinghua.edu.cn}
\and Jun He\\
Renmin University \\
{\tt\small hejun@ruc.edu.cn}
}

\begin{document}
\maketitle
\begin{abstract}

Audio-driven human gesture synthesis is a crucial task with broad applications in virtual avatars, human-computer interaction, and creative content generation. Despite notable progress, existing methods often produce gestures that are coarse, lack expressiveness, and fail to fully align with audio semantics. To address these challenges, we propose ExGes, a novel retrieval-enhanced diffusion framework with three key designs: (1) a Motion Base Construction, which builds a gesture library using training dataset; (2) a Motion Retrieval Module, employing constrative learning and momentum distillation for fine-grained reference poses retreiving; and (3) a Precision Control Module, integrating partial  masking and stochastic masking to enable flexible and fine-grained control. Experimental evaluations on BEAT2 demonstrate that ExGes reduces Fréchet Gesture Distance by 6.2\% and improves motion diversity by 5.3\% over EMAGE, with user studies revealing a 71.3\% preference for its naturalness and semantic relevance. Code will be released upon acceptance.

\end{abstract}    
\section{Introduction}
\label{sec:intro}

 Digital human animation has become a foundational technology across various domains, including virtual avatars 
 \cite{10522613}, human-computer interaction \cite {speakerco}, and creative content generation 
 \cite{Nyatsanga_2023}. Among these, audio-driven gesture synthesis—the generation of human-like gestures synchronized with input audio—plays a vital role in creating immersive and realistic virtual characters. This technology is essential for developing lifelike and expressive avatars, enabling diverse applications in entertainment, virtual communication, and interactive media \citep{alexanderson2023listen,zhu2023taming,zhi2023livelyspeaker,liu2023emage}.

\begin{figure}[t]
\centering
\includegraphics[scale=0.32]{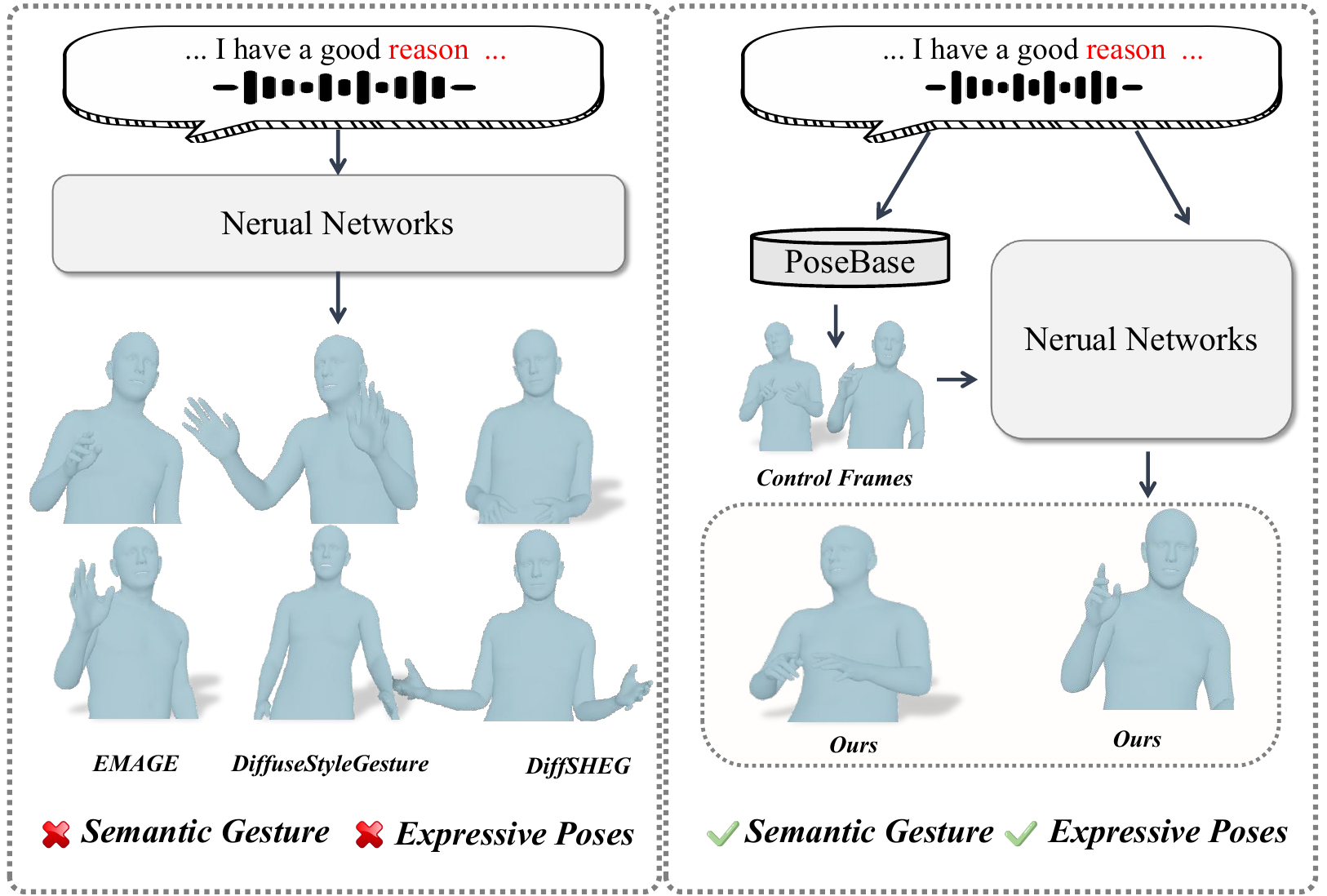}
\caption{Difference between ExGes and existing methods. ExGes aims at generating gestures that are more expressive and semantic aligning with the audio.}
\label{figure1}
\end{figure}
With the rapid advancement of audio-driven human pose animation, existing methods have made significant progress in generating realistic and synchronized gestures. VQ-VAE-based approaches \citep{van2017neural,liu2023emage,li2024lodge,chen2024diffsheg} quantize human poses using codebooks \citep{geng2023human}, enabling the generation of plausible motions but often losing fine-grained gesture details. Diffusion-based methods \citep{chen2024diffsheg,ao2023gesturediffuclip,yang2023diffusestylegesturestylizedaudiodrivencospeech}, on the other hand, produce more vivid and natural poses by leveraging regression-based strategies \citep{motionmatching}. However, as shown in Fig. \ref{figure1}, these approaches fail to generate expressive gestures that fully align with audio semantics. This limitation arises from the quantization-induced detail loss in VQ-VAE models\cite{liu2022reduceinformationlosstransformers} and the tendency of diffusion-based methods to favor commonly used gestures, restricting diversity and expressiveness. Consequently, the synthesized gestures struggle to precisely reflect speech semantics and lack the diversity needed to capture emotional states, personalized styles, and contextual adaptability.

Indeed, generating expressive and semantically aligned gestures from audio alone is inherently challenging due to the ambiguity of mapping audio features to fine-grained motion details. Given this difficulty, in this paper,  we propose that instead of relying solely on audio, external auxiliary guidance could be utilized to enhance the generation process. By incorporating such guidance through conditional control, we believe it is possible to produce gestures that are both expressive and well-aligned with speech.

To this end, we introduce ExGes, a novel retrieval-based audio-driven gesture generation framework designed to enhance expressive gestures by incorporating semantic information. The core idea behind ExGes is to address the limitation by leveraging a retrieval mechanism to provide auxiliary guidance. This approach not only enriches the expressiveness of generated gestures but also ensures their semantic alignment with the input audio.  The ExGes framework is composed of three key modules: the Motion Base Construction Module, the Motion Retrieval Module, and the Precise Control Module. The Motion Base Construction Module builds a comprehensive repository of expressive gestures, serving as a rich source of diverse and fine-grained motion patterns. Then The Motion Retrieval Module acts as a bridge between the input audio and the motion base, retrieving gestures that are semantically relevant and contextually appropriate for the given speech. Finally, the Precise Control Module integrates the retrieved gestures as conditional control signals, enabling the generation of refined and expressive motions that align naturally with the audio input. By combining these components, ExGes introduces a novel perspective on gesture generation, emphasizing the importance of auxiliary guidance and precise control to overcome the challenges of expressiveness and semantic alignment. 

Extensive experiments demonstrate that the proposed method achieves significant improvements in expressive gesture generation compared to existing approaches. The results show that the model outperforms strong baselines, such as EMAGE \cite{liu2023emage}, in overall performance, with only a negligible trade-off in certain aspects. Our contributions can be summarized as:

\begin{itemize}
\item We propose ExGes, a retrieval-based framework that leverages auxiliary guidance to improve expressive and semantically aligned human gesture generation.

\item We design three key modules: Motion Base Construction, Motion Retrieval, and Precise Control, forming an effective and scalable generation pipeline.


\item  We conduct extensive experiments, showing that ExGes significantly outperforms strong baselines, such as EMAGE \cite{liu2023emage}, with only a negligible trade-off.

\end{itemize}

\section{Related Work}
\label{sec:related}
\subsection{ Direct-regression  based Audio Driven Gesture Synchronization}
Direct-regression based methods for audio-driven gesture generation aim to directly map speech inputs to gesture outputs, addressing the complex many-to-many relationship between audio features and gestures \citep{ginosar2019learning}. Early rule-based approaches \citep{Cassell_Vilhjálmsson_Bickmore_2001,cassell1994rule} were limited in diversity and control. With advancements in deep learning, neural network-based models \citep{ginosar2019learning,Ao_Gao_Lou_Chen_Liu_2022} have improved expressiveness by learning direct mappings, though most operate in 2D, restricting their realism in practical use. To address these limitations, the introduction of 3D gesture datasets \citep{ghorbani2022zeroeggszeroshotexamplebasedgesture,yi2022generating,liu2023emage} has enabled more lifelike gesture generation. Incorporating multimodal information \citep{ao2023gesturediffuclip} and emotion-driven disentanglement \citep{chhatre2024emotionalspeechdriven3dbody} further enhances gesture diversity and contextual relevance. Recent models focus on improved control, with text-driven approaches \citep{chen2024enabling} and pose-guided methods \citep{liu2023emage} offering finer adjustments. Diffusion-based models \citep{zhong2024smoodi,chemburkar2022moddm} have also shown superior performance in generating realistic 3D gestures directly from audio. Despite these advances, challenges in fine-grained control and realism remain, especially in complex 3D environments.

While direct-regression methods outperform rule-based systems, they struggle to generate expressive and fine-grained gestures due to the unnatural ``averaging" effect caused by the many-to-many mapping between audio and gestures. To address this, we propose a retrieval-enhanced approach that incorporates example-based gesture retrieval to enhance expressiveness.

 \subsection{Retrieval based Audio Driven Gesture Generation }
In contrast to direct gesture generation approaches \citep{yang2023diffusestylegesturestylizedaudiodrivencospeech,xu2024mambatalk}, retrieval-based models \citep{yang2023QPGesture,motionmatching,ferstl2021expressgesture} encode motion data into latent embedding spaces and retrieve similar embeddings for decoding. These models often use an encoder-decoder architecture, such as VQ-VAE \citep{van2017neural}, to map retrieved embeddings to target gestures. Recent advancements focus on improving retrieval accuracy, with \citet{yang2023QPGesture} retrieving motion sequences from hidden spaces, and \citet{motionmatching} leveraging k-NN to extract latent similarities. Research also shows that standard motion embeddings enhance control and flexibility. For instance, \citet{ao2023gesturediffuclip} align motion embeddings with control information for more adaptable gesture control, while \citet{motionspace} improve semantic gesture generation by using predefined gesture candidates. Additionally, generative retrieval models like \citet{Moconvq} extend language models to generate more expressive, contextually appropriate gestures.

Different from previous works, we didn't use any extra dataset or predefined candidates. Instead, we propose a precisely control signs based method that enrich the model's expressive performance through frame-level control in this paper to improve expressiveness.

\begin{figure*}[htbp]
\centering
\includegraphics[scale=0.65]{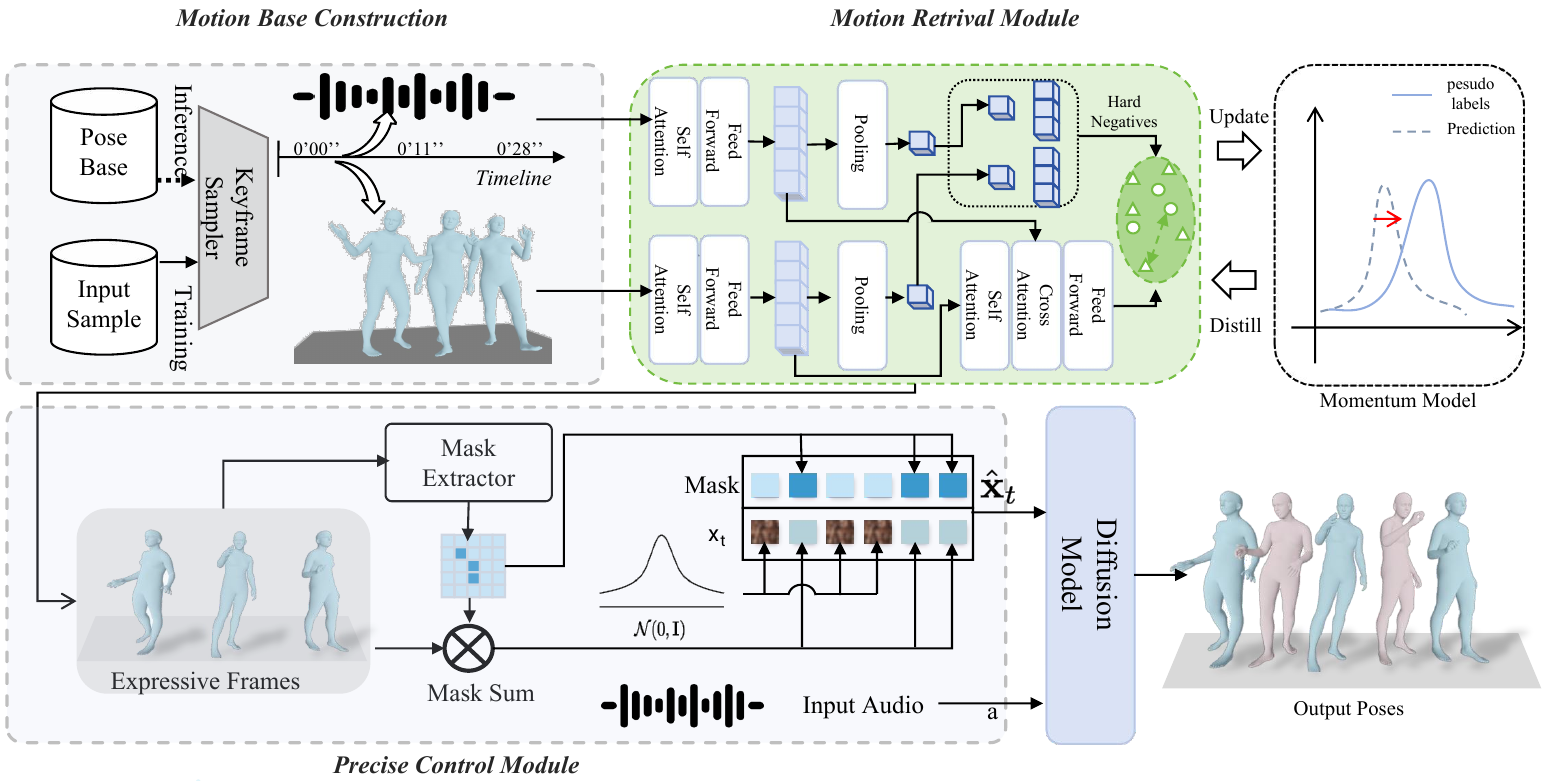}
\vspace{-0.1in}
\caption{Overview our method. The audio is first used to retrieval to get expressive gestures. The the retrieval frames are served as the control signs that insert into the Precise Control Module, enhancing the expressiveness of generated gestures.  }
\vspace{-0.1in}
\label{figure3}
\end{figure*}

\subsection{Controllable Human Motion Generation}
Controllable human motion generation focuses on synthesizing movements guided by control signals such as text descriptions \citep{lin2023text2motion}, action styles \citep{jiang2022efficient}, or trajectories \citep{kania2021trajevae}. Early methods primarily employ Variational Autoencoders (VQ-VAE) \citep{zhong2023attt2m,yi2022generating,kong2023priority} and Generative Adversarial Networks (GAN) \citep{barsoum2018hp,men2022gan,shiobara2021human}, which, while effective, have limitations in generating diverse and realistic motions. Recent advances with diffusion-based models \citep{zhong2024smoodi,chemburkar2022moddm} demonstrate superior performance, significantly improving motion realism and diversity. The introduction of CLIP \citep{radford2021learningtransferablevisualmodels} further advances text-to-motion synthesis by aligning textual inputs with corresponding motion outputs. However, these methods still struggle to provide fine-grained control over specific motion attributes such as trajectories, keyframes, and individual joints. To address these limitations, techniques such as classifier-free guidance \citep{rombach2022high} and motion inpainting \citep{Chung_Sim_Ryu_Ye_2022} have been proposed, offering more precise control over motion generation. Additionally, \citet{karunratanakul2023gmd} introduce trajectory-based control to improve motion accuracy, while \citet{cohan2024flexible} expand this control to full-body movements. Keyframe-based methods also gain traction, with \citet{li2024lodge} and \citet{ng2024audio2photoreal} demonstrating how keyframes enhance both expressiveness and the synthesis of gesture sequences.

In this paper, we build on these advancements by introducing a framework that enhances the generation of more expressive gestures with flexible control.

\section{Method}


Given an audio signal ($a \in \mathbb{R}^{T \times Q}$) and observed kinematic constraints ($c \in \mathbb{R}^{N \times J \times Q}$), our goal is to synthesize temporally coherent human motion sequences ($x \in \mathbb{R}^{N \times J \times Q}$). Here, $N$ represents the number of frames, $J$ denotes the number of joints, and $Q$ indicates the dimensionality of joint features. To enhance the expressiveness of the generated motions, we introduce an additional biomechanical constraint, $c^{*}$, which encapsulates expressive poses retrieved from a pre-built audio-motion database. The objective can be formulated as:

\begin{equation}
x = \mathcal{F}(a, c, c^{*}) = \mathcal{G}(\mathcal{H}(a), \Phi(c), \Psi(c^{*})),
\end{equation}

where $\mathcal{F}$ is the overall generative function synthesizing human motion sequences, $\mathcal{G}$ is the fusion function that combines the embeddings, $\mathcal{H}(a)$ extracts meaningful temporal and semantic audio features from the input audio $a$, $\Phi(c)$ captures spatial and temporal dependencies from the kinematic constraints $c$, and $\Psi(c^{*})$ embeds the expressive biomechanical constraint $c^{*}$, which is obtained through a semantic-aware retrieval process. This formulation ensures the generated motion $x$ is both temporally coherent and expressive.

To build the model following the above definition, as shown in Fig. \ref{figure3}, we propose ExGes, a retrieval-based controllable audio-driven human gesture generation method designed to enhance the expressiveness of generated gestures. ExGes adopts a diffusion-based architecture, where given an input audio, human gestures are synthesized through a step-by-step denoising process. Additionally, we incorporate conditional control mechanisms to ensure that the generated poses are more expressive. To achieve this, we design three key modules to obtain the conditional latent variables: (1) Motion Base Construction, (2) Motion Retrieval Module, and (3) Precise Control Module. The details of these modules are introduced as follows: Section \ref{sec:build} explains how we construct the motion base; Section \ref{sec:retrieval} introduces the retrieval-based method for generating expressive frames; finally, we present the  precise control  strategy that enables flexible control.


\subsection{Motion Base Construction}
\label{sec:build}
As shown in Fig. \ref{figure3}, to address the intricate many-to-many mapping between speech and motion for expressive gesture generation, we construct a comprehensive motion base designed to provide rich and diverse motion as auxiliary  guidance. This library is meticulously crafted to capture the nuanced interplay among linguistic, prosodic, and kinematic features, thereby enabling the generation of robust, expressive gestures. However, a critical challenge in building such a motion base lies in effectively harnessing contextual information while preserving the intricate correspondence between expressive gestures and their associated audio cues.

To overcome this challenge, we introduce a temporally aligned segmentation framework that synchronizes audio and motion data into coherent, second-level sequences. This framework integrates advanced semantic processing with adaptive temporal constraints, ensuring a highly accurate and meaningful segmentation of speech signals. Specifically, we leverage a cutting-edge automatic speech recognition (ASR) model \cite{radford2022robustspeechrecognitionlargescale} to detect linguistic boundaries at the token level, enabling precise alignment between speech segments and their corresponding motion sequences. This alignment ensures that the temporal and semantic dynamics of both modalities are faithfully preserved. To further refine the segmentation process, we also design dynamic temporal constraints that regulate segment durations to fall between 1 and 2 seconds. This design ensures that each segment encapsulates 4–8 complete words, achieving an optimal balance between preserving linguistic granularity and maintaining the continuity of motion dynamics. By avoiding excessive segmentation of single tokens, our method minimizes the risk of introducing irrelevant or misleading audio features while preserving the semantic coherence of speech and gesture sequences. 

\subsection{Motion Retrieval Module}
\label{sec:retrieval}

After constructing the expressive gesture database, the next critical step is to retrieve the most semantically and contextually appropriate gestures from this database to serve as conditional control signals. These retrieved gestures provide crucial guidance for the subsequent generation process, ensuring that the final output aligns with the intended expressive and semantic requirements. To achieve this, we propose a novel motion retrieval module that leverages a contrastive learning framework designed specifically to capture nuanced audio-motion correlations. To train this module effectively, we incorporate two key components: (1) Hard Negative Contrastive Learning and (2) Momentum Distillation. Below, we detail these components and their contributions to the retrieval process.

\noindent \textbf{Hard Negative Contrastive Learning.}  
To align audio and motion in a shared feature space, we employ dedicated audio and motion encoders that map the respective inputs into feature representations $a$ and $m$. To extract a global representation for both modalities, we apply a max-pooling mechanism. This approach assumes that the most expressive gestures and audio features can be represented by their maximum activations, resulting in pooled global embeddings $\hat{a}=P(a)$ and $\hat{m}=P(m)$. Inspired by the ALBEF framework \cite{ALBEF}, we maintain two separate queues to store the most recent $O$ audio and motion representations produced by the momentum unimodal encoders. These representations are then compared using a similarity function $s$, defined as $s(\hat{a},\hat{m})=\hat{a}^T \cdot \hat{m}$.

To compute the dual softmax similarity scores, we define the audio-to-motion and motion-to-audio similarities as follows:
\begin{equation}
\rho^{a2m}_j=\frac{\exp(s(\hat{m},\hat{a_j})/\tau)}{\sum_{j=1}^O \exp(s(\hat{m},\hat{a_j})/\tau)},
\label{a2m}
\end{equation}
\begin{equation}
\rho_j^{m2a}=\frac{\exp(s(\hat{m},\hat{a_j})/\tau)}{\sum_{j=1}^K \exp(s(\hat{m},\hat{a_j})/\tau)},
\label{m2a}
\end{equation}
where $\tau$ is a temperature hyperparameter. The ground truth one-hot similarity labels are denoted as $y^{a2m}$ and $y^{m2a}$. The contrastive learning loss function $\zeta_{c}$ is then formulated as:
\begin{equation}
\begin{split}
\zeta_{c} = \frac{1}{2} \mathbb{E}_{(a, m) \sim D}\big[ & H(y^{a2m}(a), \rho^{a2m}(a)) \\
& + H(y^{m2a}(m), \rho^{m2a}(m)) \big],
\end{split}
\end{equation}
where $H$ is the cross-entropy loss function.

To further enhance the discriminative power of the learned representations, we incorporate a binary classification task during contrastive learning. A fully connected layer predicts whether an audio-motion pair is semantically aligned, with the binary classification loss defined as:
\begin{equation}
\zeta_{m}=\mathbb{E}_{(a, m) \sim D} H\left(y^{\mathrm{itm}}, p^{\mathrm{itm}}(a,m)\right),
\label{equ.matching}
\end{equation}
where $y^{itm}=1$ if the audio $a$ and motion $m$ are aligned, and $y^{itm}=0$ otherwise.

Given the many-to-many mapping relationship between audio and motion, conventional negative sampling strategies often fail to capture the diversity of potential matches, which can hinder the effectiveness of contrastive learning. To address this, we propose a hard negative sampling strategy based on the contrastive similarity defined in Equations \ref{a2m} and \ref{m2a}. For each audio instance in a mini-batch, we sample a negative motion sequence with high similarity from the same batch as a hard negative. Similarly, for each motion sequence, we sample a hard negative audio instance. This bidirectional sampling strategy ensures diversity while strengthening the discriminative capacity of the model.

\noindent  \textbf{Momentum Distillation.}  
Despite the use of exactly matched audio-motion pairs in our dataset, the inherent many-to-many mapping relationship can degrade the reliability of the learned representations. To mitigate this issue, we employ momentum distillation \cite{ALBEF}, a teacher-student framework, where a momentum-based teacher model generates pseudo-labels to guide the training of the base model. During training, the teacher model computes audio-motion similarity scores using features $\hat{a}'$ and $\hat{m}'$ from the momentum unimodal encoders. These scores are used to derive soft pseudo-targets $y^{a2m}$ and $y^{m2a}$, which replace the original hard labels in Equations \ref{a2m} and \ref{m2a}. The momentum distillation loss is defined as:
\begin{equation}
\begin{split}
\zeta_{\text{ITM}} = \mathbb{E}_{(a, m) \sim D} \big[ & \operatorname{KL}\left(\boldsymbol{y}^{a2m}(a) \| \rho^{a2m}(a)\right) \\
& + \operatorname{KL}\left(\boldsymbol{y}^{m2a}(m) \| \rho^{m2a}(m)\right) \big],
\end{split}
\label{momentum}
\end{equation}
where $\operatorname{KL}$ denotes the Kullback-Leibler divergence.

The total loss function for training the motion retrieval module is then defined as:
\begin{equation}
\zeta = \alpha \zeta_{\text{ITM}} + (1 - \alpha) \zeta_{m},
\end{equation}
where $\alpha$ is a hyperparameter controlling the trade-off between the two loss components.

\noindent \textbf{Expressive Gesture Retrieval.}  Once the retrieval model is trained, it allows us to identify expressive keyframes/gestures and their temporal locations from the motion database. By leveraging the max-pooling mechanism used in feature extraction, the most expressive keyframe can be directly identified by finding the frame with the maximum similarity score, formulated as $l = \arg \max(s(a, m))$. The corresponding pose at location $l$ serves as the keyframe $c$, which acts as a critical control signal for guiding the generation of semantically aligned and expressive gestures.

\begin{figure}[htbp]
\centering
\includegraphics[scale=0.4]{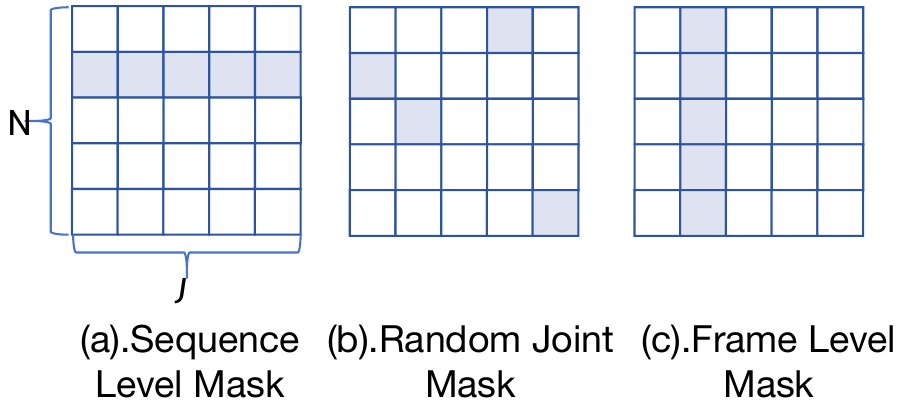}
\caption{Different mask type in our work. The {\color{blue}blue} squares indicates the control information and the white information indicates the noise location.}
\label{fig.mask}
\end{figure}

\subsection{Precision Control Module}
\label{control}
After retrieving the expressive keyframes/gestures , another critical challenge lies in effectively incorporating the control gestures into the generation process. Unlike traditional image-based inpainting methods \cite{lugmayr2022repaintinpaintingusingdenoising}, integrating single-frame gestures poses unique difficulties, as they may become indistinguishable from noise within the same distribution. Conversely, incorporating multiple frames risks disrupting the temporal beat alignment, which is crucial for maintaining natural motion dynamics. To tackle this problem, we propose a novel approach for seamlessly applying frame-level conditions during the diffusion-based generation process.

\noindent  \textbf{Motion Control with Partial Feature Masking.}  
To explicitly decouple the expressive pose ($\mathbf{c}$) from the noise-corrupted input ($\mathbf{x}_t$), we propose a binary mask $m \in \mathbb{R}^{N \times W \times Q}$ where $N$, $W$, and $Q$ respectively denote the batch size, joint count, and pose feature dimension. This mask serves as a selection mechanism to distinguish noise components from valid pose features. The denoised input $\hat{\mathbf{x}}_t$ is computed through element-wise composition: 
\begin{equation}
\hat{\mathbf{x}}_t = m \odot \mathbf{x}_0 + (1-m) \odot \mathbf{x}_t
\end{equation}
where $m \odot \mathbf{x}_0$ preserves observed pose information while $(1-m) \odot \mathbf{x}_t$ retains noise components. Unlike EMAGE\cite{liu2023emage} that applies masking to entire features, our partial masking strategy preserves partial structural information while enabling three distinct control mechanisms: \textit{1) Trajectory-based} control for global motion paths, \textit{2) Joint-level} control for anatomical precision, and \textit{3) Frame-wise} control for temporal dynamics (see Fig.\ref{fig.mask}). This hierarchical design enhances both motion controllability and generation accuracy.

\noindent  \textbf{Modality Balancing via Stochastic Masking.}  
We empirically observe the model's tendency to overfit to audio cues while underutilizing numerical pose data. To address this imbalance, we implement  stochastic masking: during training, we randomly discard key frames and audio inputs with probability $\phi$.  In addition, we use curriculum learning during training of our model. We decrease the mask rate from 90\% to 3\% during training process to make it easy for the Precise Control Module to learn the control information.

\section{Experiments}

\section{Implementation Details}
Our model is built upon the DiffuseStyleGesture framework \citep{yang2023diffusestylegesturestylizedaudiodrivencospeech} and is trained for 200,000 steps on a single A6000 GPU, with the entire process taking approximately 24 hours. During training, we apply both masking and noise control to 70\% of the input samples to improve robustness. Specifically, random keypoint masks are applied to 40\% of the samples, frame-level masks to 30\%, and keypoint sequence-level masks to the remaining 30\%. To prevent overfitting to specific control signals, we randomly set either the audio or control signals to null for 10\% of the inputs. Furthermore, we adopt a curriculum learning strategy, gradually decreasing the mask rate from 90\% to 3\% over the course of training. Our experiments are conducted on the {BEAT2} dataset, a large-scale dataset containing 27 hours of speech data from 30 speakers. We follow the standard protocol used in previous works \citep{liu2023emage}, splitting the dataset into 85\% for training, 7.5\% for validation, and 7.5\% for testing.

\subsection{Metrics}
To evaluate the effectiveness of our method, we assess the model's ability to generate expressive gestures from two key aspects: overall generation quality and the performance of the Precision Control Module.  For generation quality, we utilize the following metrics: Fréchet Gesture Distance (FGD) \citep{liu2022disco}, Beat Consistency (BC) \citep{AIST++}, and Diversity (Div) \citep{li2021audio2gestures}. These metrics respectively measure the accuracy, audio alignment, and variability of the generated gestures. For control quality, we employ human pose estimation metrics, including Mean Per Joint Position Error (MPJPE) and Procrustes Aligned MPJPE (PA-MPJPE), to evaluate the fidelity and precision of the gestures.  As it is challenging to quantitatively evaluate the Motion Retrieval Module, we provide qualitative results instead to demonstrate its effectiveness.


\begin{table}
    \centering
    \small
    \begin{tabular}{ccccc}
        \toprule
        Method & FGD$\downarrow$ & BC$\uparrow$ & Diversity$\uparrow$&\makecell{Fine-Grained  \\Control\\} \\
        \midrule
        S2G\cite{ginosar2019learning} & 28.15 & 4.83 & 5.971 &\XSolidBrush\\
        Trimodal\cite{yoon2020speech} & 12.41 & 5.933 & 7.724 &\XSolidBrush\\
        HA2G\cite{liu2022learning} & 12.32 & 6.779 & 8.626&\XSolidBrush \\
        DisCo\cite{liu2022disco} & 9.417 & 6.439 & 9.912 &\XSolidBrush\\
        CaMN\cite{liu2022beat} & 6.644 & 6.769 & 10.86 &\XSolidBrush\\
        \makecell{DiffuseStyle\\Gesture\cite{yang2023diffusestylegesturestylizedaudiodrivencospeech}} & 8.811 & 7.241 & 11.49 &\XSolidBrush\\
        Habibie\cite{habibie2021learning} & 9.04 & 7.716 & 8.231 &\XSolidBrush\\
        Talkshow\cite{yi2022generating} & 6.209 & 6.947 & \underline{13.47}&\XSolidBrush \\
        EMAGE\cite{liu2023emage} & \underline{5.512} & \textbf{7.724} & 13.06&\XSolidBrush \\
              DiffSHEG\cite{chen2024diffsheg} & 10.51 & 5.55 & 10.91&\XSolidBrush \\
        \midrule
        ExGes & \textbf{5.261} &\underline{6.97} & \textbf{13.75} &\Checkmark\\
        \bottomrule
    \end{tabular}

    \footnotesize
        \caption{Performance comparison of different methods on gesture synchronization task.The \textbf{bloded} is the best result and the \underline{underline} result is the second best. }
        
        \label{tab:gesture-synchronization-performance}
\vspace{-0.1 in}
\end{table}

\begin{figure*}[htbp]
\centering
\includegraphics[scale=0.51]{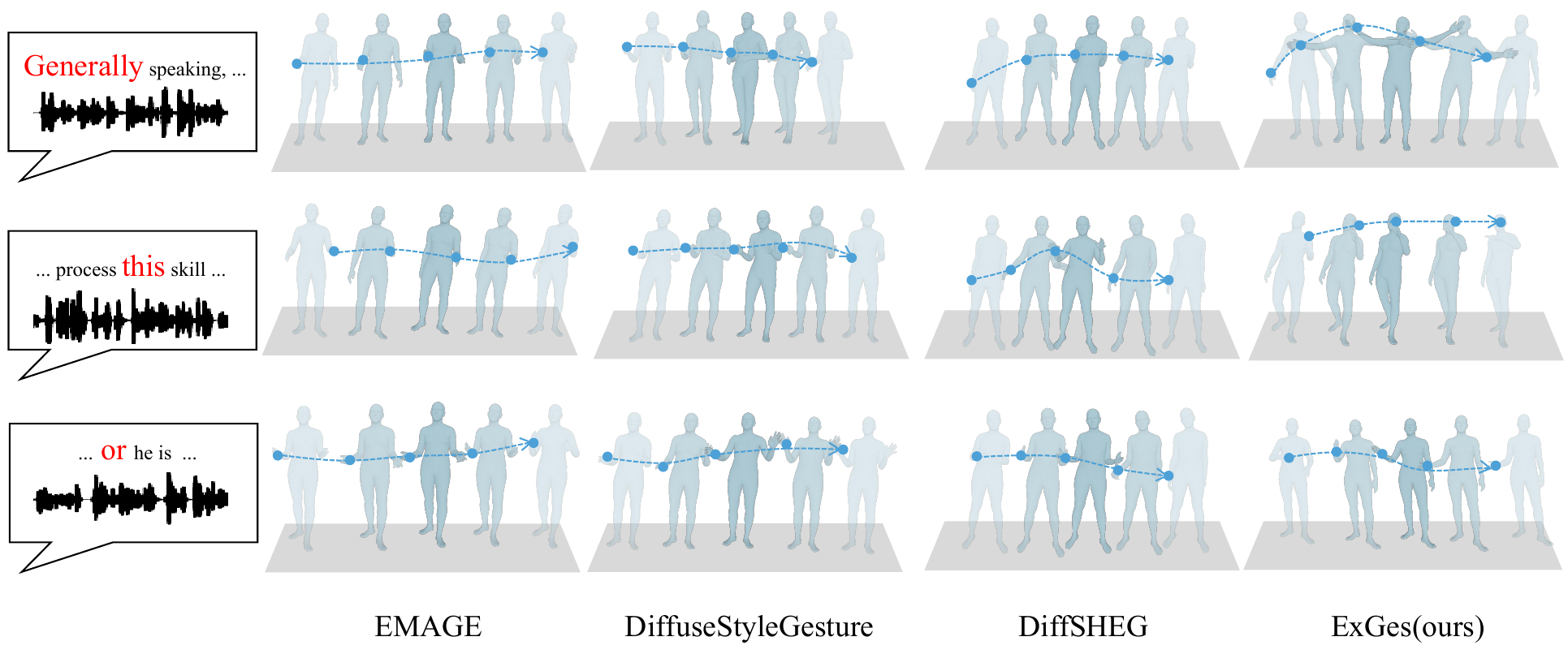}
\caption{Comparison results between EX-Ges with existing methods. We compare the outputs generated by different models under the same speech input and plotted the hand motion sequences with the richest semantic information.}
\label{fig.case}
\label{figure}
\end{figure*}

\subsection{Quantitative Results Analysis}
This section presents the experimental results on the BEAT2 dataset, where our proposed model, ExGes, achieves strong performance in both generation quality and control precision, both of which contribute to the evaluation of its expressive fine-grained gesture generation capabilities.

\noindent \textbf{Performance on Generation Related  Metrics.}  As shown in Table \ref{tab:gesture-synchronization-performance}, our method achieves the lowest Fréchet Gesture Distance (FGD) (5.261), surpassing all existing approaches, including DiffuseStyleGesture and Talkshow, with reductions of 2.69 and 0.088, respectively. This indicates that our model generates gestures that are more closely aligned with the ground truth, reflecting superior accuracy in motion generation. Additionally, ExGes leads in Diversity, achieving a score of 13.75, which is not only the highest among all compared methods but also a 0.26 improvement over EMAGE, demonstrating the model's capability to produce a wide range of varied and natural gestures. While our model shows a slight decrease in Beat Consistency compared to DiffuseStyleGesture (6.97 vs. 7.241), it remains highly competitive, ranking second-best in this metric. This suggests that our model maintains strong synchronization with the audio, even as it excels in other critical areas such as gesture accuracy and diversity.

   \begin{table}
   \small
    \centering
    \scalebox{0.9}{
    \begin{tabular}{cccc}
        \toprule
        Method & Frame & PAMPJPE (mm) $\downarrow$ & MPJPE (mm) $\downarrow$ \\
        \midrule
        \multirow{3}{*}{\makecell{DiffuseStyle-\\Gesture}} & 1 & 84.3 & 119.2 \\
        & 2 & 86.4 & 114.4 \\
        & 3 & 85.76 & 114.1 \\
        \midrule
        \multirow{3}{*}{EMAGE} & 1 & 83.9 & 108.6 \\
        & 2 & 80.8 & 114.0 \\
        & 3 & 71.7 & 112.8 \\
        \midrule
        \multirow{3}{*}{ExGes} & 1 & 66.1 & 103.5 \\
        & 2 & 47.2 & 86.5 \\
        & 3 & \textbf{39.4} & \textbf{75.1} \\
        \bottomrule
    \end{tabular}
    }
    \caption{Control Effectiveness Evaluation for Full-Body Human Motion. This table shows the PAMPJPE and MPJPE values (in mm) for different methods when inserting 1, 2, or 3 control frames. Lower values indicate better control accuracy and higher quality of the generated gestures.}
    \label{tab:control-effectiveness}
    \vspace{-0.1in}

\end{table}

\noindent \textbf{Performance on Control Related Metrics.} To evaluate our model's precision control capability, we conduct a systematic comparison with state-of-the-art controllable gesture synthesis methods: DiffuseStyleGesture\cite{yang2023diffusestylegesturestylizedaudiodrivencospeech} using its proposed inpainting-based control mechanism, and EMAGE\cite{liu2023emage} utilizing its skeletal conditioning strategy. We quantitatively assess control precision through progressively challenging scenarios with 1- to 3-frame control inputs. As evidenced by Table \ref{tab:control-effectiveness}, our method achieves substantial improvements in pose accuracy across all settings - reducing MPJPE by 21.2\% (83.9→66.1mm) under single-frame control compared to EMAGE, and delivering 46.3mm lower PA-MPJPE than DiffuseStyleGesture in tri-frame control scenarios. These results demonstrate our framework's superior ability to propagate sparse control signals while maintaining motion naturalness, particularly in challenging multi-frame coordination tasks.

\subsection{Qualitative Results Analysis}
We evaluate the performance of our model across three key aspects: speech generation quality, action control effectiveness, and action retrieval performance, comparing our approach against three baseline methods—EMAGE, DiffSHEG, and DiffuseStyleGesture. Specifically, we compare the outputs generated by different models under the same speech input to emphasize the higher quality and richer semantics of the gestures produced by our model.  To evaluate retrieval performance, we present the top three results for speech retrieval from the test set. For action control, we demonstrate the effectiveness of our model in controlling different body parts. Detailed comparison results and ablation studies for the control module are provided in the appendix.
\begin{figure}[t]
\centering
\includegraphics[scale=0.4]{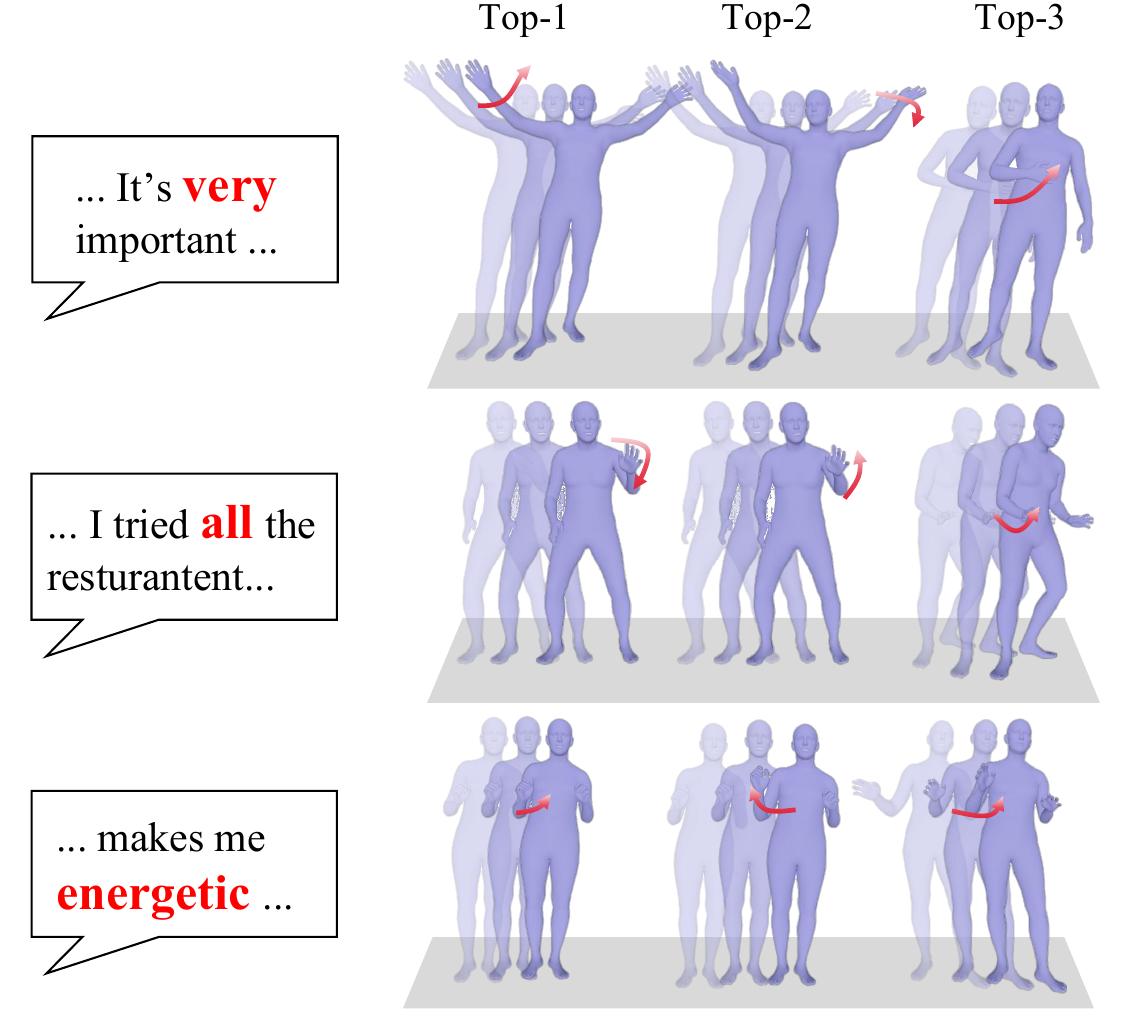}
\caption{Visualization  results of of the searched keyframes.}
\vspace{-0.2in}
\label{fig.search}
\end{figure}

As shown in Fig.\ref{fig.case}, our model demonstrates a clear advantage in capturing expressive human motion details that naturally align with the spoken content. Compared to DiffuseStyleGesture, DiffSHEG, and EMAGE, our approach generates gestures that are more semantically relevant to the speech.  For instance, in the first row, when generating a gesture for the word "generally," our model effectively retrieves and controls a gesture where the hands spread open, emphasizing the speech's intent. Likewise, for demonstrative words like "this," our method accurately produces a pointing gesture with the index finger, highlighting emphasis. Additionally, for the word "or," the generated gesture extends outward in a circular motion, underscoring the semantic weight of the word. 

Next, we examine the performance of our action retrieval module, which plays a crucial role in generating contextually appropriate gestures based on the input speech. As illustrated in Fig.\ref{fig.search}, our model leverages contrastive learning between short speech segments and their corresponding gestures, effectively capturing the intricate relationship between speech and motion. This strong coupling enables the model to retrieve gestures that align closely with the content and intent of the speech.  For example, when the word "very" is spoken, our model retrieves a gesture where the speaker raises their hands slightly, emphasizing the intensity of the adjective. Similarly, when describing something as "energetic," the retrieved gestures are dynamic and expressive, effectively reflecting the sentiment of the speech.

\begin{figure}[t]
\centering
\includegraphics[scale=0.4]{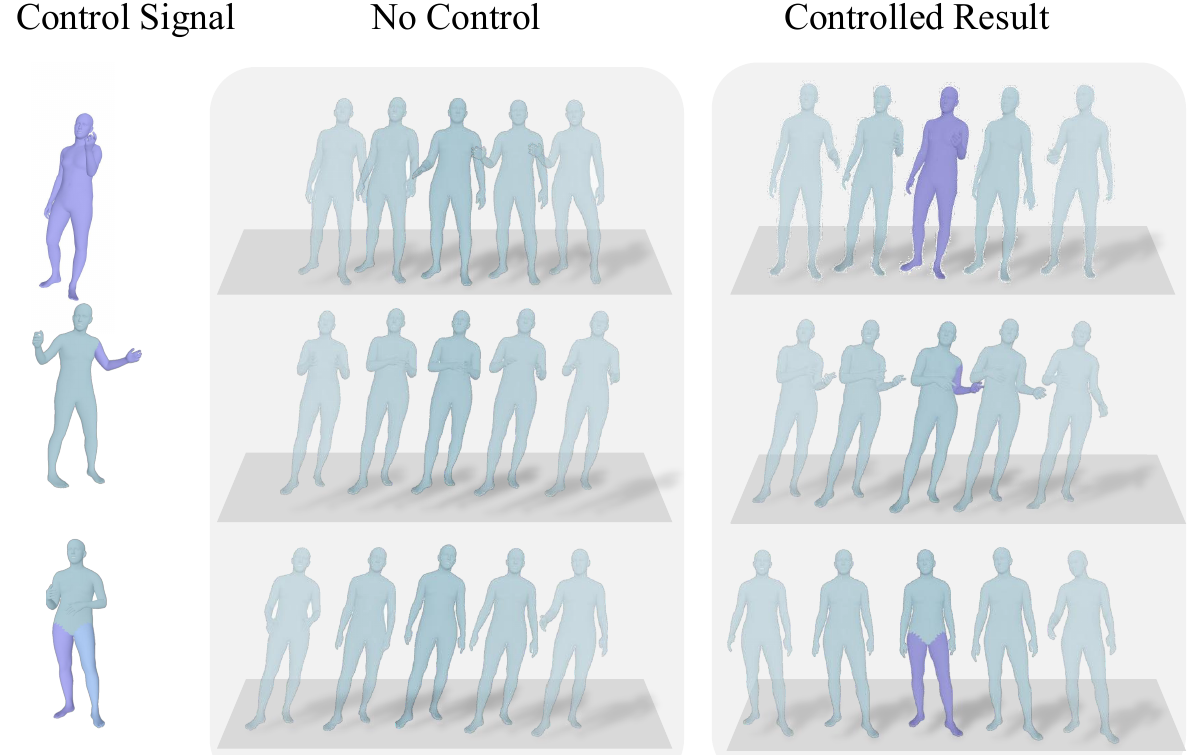}
\vspace{-0.1in}
\caption{Controllability comparison results.}
\label{control}
\vspace{-0.25in}
\end{figure}

We further evaluate the action controllability of our model through pose manipulation, as illustrated in Fig. \ref{control}. The results demonstrate three distinct control capabilities: (1) full-body pose control using single-frame reference poses, (2) directional adjustment of arm movements, and (3) modification of standing postures. Specifically, the model effectively reproduces emphasized gestures when guided by similar keyframe references, while arm trajectories and stance configurations can be precisely adjusted based on corresponding single-frame instructions. These findings underscore the model's flexibility in generating controlled and contextually accurate gestures.


\subsection{Ablation Study}

This section presents the ablation study to evaluate the contributions of the Motion Retrieval Module and the Precise Control Module (PCM) to the overall performance of our model. As summarized in Table \ref{tab.ablation}, removing either module results in a noticeable decline in performance, underscoring their critical importance.

\noindent \textbf{Impact of the Motion Retrieval Module:}  
As shown in Table \ref{tab.ablation}, when this module is removed, the model maintains a comparable level of diversity (13.22); however, it suffers a significant loss in precision, as indicated by an increase in Fréchet Gesture Distance from 5.22 to 6.50. This suggests that while this module is not integral to maintaining diversity, it plays a crucial role in enhancing the precision and accuracy of the generated gestures.

\noindent \textbf{Impact of the Precise Control Module:}  
As shown in Table \ref{tab.ablation}, the absence of this module leads to a sharp performance decline across all metrics. FGD increases substantially to 7.80, while Beat Consistency and Diversity drop to 6.70 and 11.49, respectively. This highlights the PCM’s vital role in capturing expressive motion details and ensuring robust alignment between audio and gesture. Furthermore, the ablation study on part mask removal reveals consistent performance degradation: Diversity decreases from 13.75 to 13.48, BC drops from 6.97 to 6.89, and FGD rises from 5.22 to 5.40. These results demonstrate that part masks are essential for improving the model’s ability.

\noindent \textbf{Full Model Integration:}  
As shown in Table \ref{tab.ablation}, the complete integration of all modules in the ExGes framework achieves the best results, with an FGD of 5.22, BC of 6.97, and Diversity of 13.75. These findings confirm that the combination of EGRM, PCM, and the diffusion module significantly enhances both the precision and diversity of the generated gestures, leading to superior performance over strong baselines across all key metrics.

\begin{table}[htbp]
\centering

\label{tab:comparison}
\begin{tabular}{cccc}
\toprule
Method & FGD$\downarrow$ & BC$\uparrow$ & Diversity$\uparrow$ \\
\midrule
\makecell{w/o Motion Retrieval }      & 6.5 & 6.90 & 13.22 \\
\makecell{w/o Precision Control} & 7.80 & 6.70 & 11.49 \\
\makecell{w/o Part Mask}& 5.4&6.89&13.48\\
\midrule
ExGes                          & \textbf{5.22} & \textbf{6.97} & \textbf{13.75} \\
\bottomrule
\end{tabular}

\caption{Ablation study result of our model.}
\label{tab.ablation}
\end{table}
\begin{table}
\centering
\scalebox{0.85}{
\begin{tabular}{cccc}
\toprule
Comparison & Metric & Baseline (\%) & Ours (\%) \\
\midrule
\multirow{ 3}{*}{Ours vs. EMAGE} & Liveness & 14.30\% & \textbf{85.60\%} \\
               & BC & 22.50\% & \textbf{77.50\%} \\
            ~& Expressiveness &17.5\%&\textbf{82.5\%}\\
\midrule
\multirow{ 3}{*}{\makecell{Ours vs. \\DiffuseStyleGesture }}& Liveness & 25.50\% & \textbf{74.50\%} \\
   ~                          & BC & 19.60\% & \textbf{80.40\%} \\
~&  Expressiveness &24.3\%&\textbf{75.7\%}\\
\midrule
\multirow{ 3}{*}{\makecell{Ours vs. \\DiffSHEG }}& Liveness & 9.20\% & \textbf{90.80\%} \\
   ~                          & BC & 21.70\% & \textbf{78.30\%} \\
~&  Expressiveness &4.3\%&\textbf{95.7\%}\\
\bottomrule
\end{tabular}
}
\caption{User Study Results}
\label{tab:userstudy}
\vspace{-0.75cm}
\end{table}



\subsection{User Study}
To further evaluate the effectiveness of our method, we perform a two-alternative forced choice (2AFC) test with 51 randomly selected participants. Each participant views 10 sets of results generated by ExGes, DiffuseStyleGesture, and EMAGE from the BEAT2 dataset and selects the segment that best aligns with the given speech. The results, presented in Table \ref{tab:userstudy}, show that our method achieves state-of-the-art performance compared to previous approaches, particularly excelling in both Liveness and Beat Consistency metrics. Compared to EMAGE, our method increases realism by 71.3\%, beat consistency by 55\%, and expressiveness by 65\%, demonstrating significantly improved perceptual quality. Against the diffusion-based DiffuseStyleGesture, our approach achieves a 29\% increase in realism, a 59.8\% improvement in beat consistency, and a 51.4\% boost in expressiveness. Furthermore, when compared to DiffSHEG, our method substantially improves liveness from 9.2\% to 90.8\% and enhances expressiveness by 91.4\%.

\section{Conclusion}
In this paper, we tackle the challenge of expressive audio-driven human gesture synthesis. To address the limitations of existing methods, which often generate coarse, inexpressive, and semantically misaligned gestures, we introduce ExGes, a novel retrieval-enhanced diffusion framework. ExGes integrates three key components: Motion Base Construction, a Motion Retrieval Module, and a Precision Control Module, enabling the generation of high-quality, expressive, and contextually aligned gestures. Experimental results consistently demonstrate the superiority of ExGes over existing approaches.
{
    \small
    \bibliographystyle{ieeenat_fullname}
    \bibliography{main}
}

\end{document}